\documentclass[conference]{IEEEtran}
\IEEEoverridecommandlockouts

\usepackage{cite}
\usepackage{amsmath,amssymb,amsfonts}
\usepackage{algorithmic}
\usepackage{algorithm2e}
\usepackage{graphicx}
\usepackage{textcomp}
\usepackage{xcolor}
\def\BibTeX{{\rm B\kern-.05em{\sc i\kern-.025em b}\kern-.08em
    T\kern-.1667em\lower.7ex\hbox{E}\kern-.125emX}}
\begin{document}

\title{Enhanced Multi-model Online \\ Conformal Prediction\\
\thanks{Work in the paper is supported by NSF EECS 2207457, NSF ECCS 2412484 and NSF ECCS 2442964.}
}

\author{\IEEEauthorblockN{Erfan Hajihashemi}
\IEEEauthorblockA{\textit{Department of EECS} \\
\textit{University of California, Irvine}\\
Irvine, USA \\
ehajihas@uci.edu}
\and

\IEEEauthorblockN{ Yanning Shen}
\IEEEauthorblockA{\textit{Department of EECS} \\
\textit{University of California, Irvine}\\
Irvine, USA \\
yannings@uci.edu}
}

\maketitle

\begin{abstract}
Conformal prediction is a framework for uncertainty quantification that constructs prediction sets for previously unseen data, guaranteeing coverage of the true label with a specified probability. However, the efficiency of these prediction sets, measured by their size, depends on the choice of the underlying learning model. Relying on a single fixed model may lead to suboptimal performance in online environments, as a single model may not consistently perform well across all time steps. To mitigate this, prior work has explored selecting a model from a set of candidates. However, this approach becomes computationally expensive as the number of candidate models increases. Moreover, poorly performing models in the set may also hinder the effectiveness. To tackle this challenge, this work develops a novel multi-model online conformal prediction algorithm that reduces computational complexity and improves prediction efficiency. At each time step, a bipartite graph is generated to identify a subset of effective models, from which a model is selected to construct the prediction set. Experiments demonstrate that our method outperforms existing multi-model conformal prediction techniques in terms of both prediction set size and computational efficiency.
\end{abstract}


\section{Introduction}
Most machine learning algorithm designs aim to enhance the accuracy of label prediction. Nevertheless, a
significant challenge persists as many models demonstrate limitations in predicting labels with high
certainty, falling short of achieving the desired levels of accuracy and other critical evaluation metrics. Conformal prediction methods can significantly increase the reliability of machine learning techniques by providing prediction sets that contain the true class with a predetermined probability \cite{b13,b14,b15}. However, classical CP methods assume that calibration samples are exchangeable, an assumption that is often not satisfied in online settings. To cope with distribution shifts and the resulting breakdown of exchangeability, many extensions of CP have been proposed \cite{b16,b17,b18,b19}. Despite these advancements, the efficiency (e.g., prediction set size or regret) of previous conformal prediction methods in online settings with distribution shifts heavily depends on the model employed, and a single model may not consistently perform well across various distribution shifts. Creating an
efficient prediction set size is as important as obtaining the desired coverage. To address this issue, \cite{b6,b23,b24} proposed leveraging multiple learning models to provide diverse candidates for adaptive conformal prediction algorithms to select the appropriate model. However, their approaches are limited to a set of candidate models with good performance. Moreover, multi-model conformal prediction can become computationally expensive when the number of candidate models is large, since the adaptive conformal prediction parameters must be updated separately for each model. \par

Graph-structured representations, whether directed or undirected, can be leveraged to model a wide range of real-world dependencies and relational structures \cite{b28,b29}. Recent work further exploits graph-based feedback to enhance online learning \cite{b30,b31,b32}. In this work, we introduce a new multi-model online conformal prediction method that views online model selection through a graph-structured feedback framework. The proposed method dynamically selects a subset of effective learning models and prunes weak ones at each time step by constructing a graph. \par

\subsection{Related Work}
Conformal prediction \cite{b13,b14} is an effective method for uncertainty quantification that has been widely used to predict a set of candidate labels for income data. It treats the learning model as a black box and provides a prediction set for new test data. Conformal prediction frameworks can be utilized on both classification \cite{b20} and regression \cite{b21} tasks. Conformal prediction algorithms can be broadly categorized into split and full variants \cite{b17}. In split conformal prediction, the training data is divided into two disjoint subsets: a proper training set and a calibration set. The proper training set is used to fit the point prediction model, while the calibration set is used to compute nonconformity scores \cite{b22}. In contrast, full conformal prediction is significantly more computationally demanding, as it requires retraining or scoring the point prediction model for each test point and every possible candidate label \cite{b5}. Hence, in this work, we only focus on split conformal prediction algorithms. \par
Employing standard conformal prediction in online environments, where the exchangeability assumption may be violated, does not achieve the desired coverage guarantee. To address this, \cite{b18} introduced the use of a time-varying miscoverage probability. However, this approach has certain limitations (e.g., the need to specify the learning rate in advance). \cite{b19,b25} use expert learning techniques to mitigate these limitations. \cite{b26} utilized reweighting techniques to cope with changes in online settings. However, these methods often rely on some distributional assumptions.  Despite achieving valid coverage guarantees, these methods may fail to construct efficient prediction sets that are both small and able to cover the true label. Some recent works propose using multiple learning models to enhance conformal prediction. \cite{b27} leverages multiple models in the full conformal prediction setting, which suffers from high computational cost. \cite{b24} proposes a majority-vote strategy for aggregating conformal sets in the split conformal prediction setting. However, their method suffers from coverage loss and lacks a theoretical guarantee of achieving the desired \(1-\alpha \) coverage. \cite{b6} proposed selecting a model from a set of candidate models at each time step. This approach can incur high computational cost and reduced efficiency when the set includes poorly performing candidates.

\section{Preliminaries}
In online settings, data arrive sequentially over time.  At time step \(t\), historical observation are denoted by  \(\{(X_{\tau}, Y_{\tau}^{\text{true}})\}_{\tau=1}^{t-1}\), where \(X_{\tau} \in \mathcal{X}\) represents the input data at time \(\tau\) and \(Y_{\tau}^{true} \in \mathcal{Y}\) is the corresponding true label. The test inputs \(\{X_{\tau}\}_{\tau=t}^{T}\) are then revealed one at a time. In online conformal prediction, given miscoverage probability \(\alpha\), learning model \(m\), historical data , the objective is to construct a prediction set \(C_{\alpha_{\tau}}^{m} (X_{\tau}) \subseteq \mathcal{Y}\)   for every test data \(X_{\tau}\) such that:
\begin{equation}
  \frac{1}{T-t+1} \sum_{\tau = t}^T \mathbb{I}[Y_{\tau}^{true} \in C_{\alpha_{\tau}}^m(X_{\tau})] \ge 1 - \alpha.
\end{equation}

In the online scenario, a time-variant miscoverage probability \(\alpha_{\tau}\) is employed to ensure that the true label is included in the prediction set with the desired coverage probability of \(1 - \alpha\).

Note that the decision of which candidate labels to include in the prediction set is determined by a threshold computed from non-conformity scores of historical data. These scores measure the disagreement between the true label and the model’s prediction. Utilizing non-conformity scores of historical data at time \(t\), the threshold is obtained by:
\begin{equation}
  \hat{q}_{\alpha_t}^{m} = Quantile \left(\frac{\lceil t (1-\alpha_t) \rceil}{t-1},\{S^{m}(X_{\tau},Y_{\tau}^{true})\}_{\tau=1}^{t-1}\right),
  \label{eq:quantile_formula}
\end{equation}
where \(Quantile(\cdot, \cdot)\) sorts the nonconformity scores in ascending order and then outputs \(\frac{\lceil t (1-\alpha) \rceil}{t-1}\) empirical quantile of sorted scores. Next, prediction set for new data \(X_t\) is constructed as 
\begin{equation}
    \label{eq:Create_pred_set}
    C_{\alpha_t}^{m} (X_t) = \{Y \in \mathcal{Y} \mid S^{m}(X_t, Y) \le \hat{q}_{\alpha_t}^{m} \}.
\end{equation}
For simplicity of notation, we index the test-time steps as \(t=1\) to \(T\) in the remainder of this paper. To construct more efficient prediction sets in online environments, \cite{b6} proposed leveraging multiple learning models. At each time step \(t\), the model \(\hat{m}\) is selected based on performance of every learning model \(m \in [M]\) over previous time steps. In practice, however, \(M\) candidate learning models may include poorly performing models, which can result in unnecessarily large prediction sets. Moreover, employing a large number of learning models increases the computational complexity. To mitigate these issues, we introduce a data-driven approach that selects a subset of models at each time step and uses only these selected models as candidates when constructing the prediction set.

\section{Methodology}
\subsection{Model Selection}
This section introduces a graph-aided framework for multi-model conformal prediction that addresses the challenges of computational complexity and inefficiency. At time step \(t\), consider a bipartite graph \(G_t\) \cite{b1} which includes two sets of nodes: \(M\) model nodes \(\{v_1^{(l)}, ...,v_M^{(l)} \}\) and \(J\) selective nodes \(\{v_1^{(s)}, ...,v_J^{(s)} \}\) where \(v_m^{(l)}\) and \(v_j^{(s)}\) represents \(m-\)th learning model and \(j\)-th selective node respectively.  In this work, each selective node is connected to at most \(N\) model nodes. Given \(G_t\) at each time step \(t\), one selective node is chosen and its associated model nodes, forming a subset denoted by \(S_t\), are used as the candidate set for the conformal prediction task. These selected model nodes can contribute to the conformal prediction task by selecting a single model according to a probability mass function (PMF). The selection is guided by the weight \(w_t^m\) assigned to each model \(m \in [M]\), which influences both the generation of the graph \(G_t\) and the selection of model \(\hat{m}\) from the subset of effective models \(S_t\).

After constructing the prediction set for \(X_t\), the true label \(Y_t^{true}\) is revealed and used to update the adaptive miscoverage probability. Given that each model generates a distinct prediction set, applying and updating a single miscoverage parameter \(\alpha_t\) for all models is inadequate. Instead, a specific time-variant misscovrage probability is assigned to each model \(m \in [M]\), denoted as \(\alpha_t^m\). 

To update miscoverage probability \(\alpha_t^m\) for each \( m \in [S_t]\),  we adopt the pinball loss defined as \cite{b2}:
        \begin{equation}
         L(\Bar{\alpha}_t^m, \alpha_t^m) = \alpha (\Bar{\alpha}_t^m - \alpha_t^m) - \min \{0 , \Bar{\alpha}_t^m - \alpha_t^m\},
         \label{eq:loss-model-m}
         \end{equation}
         where 
         \begin{equation}
         \Bar{\alpha}_t^m  := \sup\{\Tilde{\alpha} : Y_t^{true} \in C_{\Tilde{\alpha}}^m (X_t)\}
        \label{eq:optimum-alpha}
        \end{equation}
is the best possible value of miscoverage probability for model $m$ at time $t$, which constructs the smallest prediction set that covers $Y_t^{true}$. The miscoverage probability $\alpha_{t+1}^m$ can be updated via scale free online gradient descent (SF-OGD)  \cite{b3} as
\begin{equation}
\alpha_{t+1}^{m} = \alpha_t^m - \eta \frac{\nabla_{\alpha_t^m} L(\Bar{\alpha}_t^m, \alpha_t^m)}{\sqrt{ \sum_{\tau=1}^{t} \|\nabla_{\alpha_{\tau}^m} L(\Bar{\alpha}_{\tau}^m, \alpha_{\tau}^m)\|_2^2 }},
\label{eq:update-rule}
\end{equation}
which follows an online gradient descent update with a time-dependent decaying learning rate. The parameter $\eta$ is the learning rate and
\begin{equation}
\nabla_{\alpha_t^m} L(\Bar{\alpha}_t^m, \alpha_t^m) = \mathbb{I} [\Bar{\alpha}_t^m < \alpha_t^m ] - \alpha = err_t^{m} - \alpha, \label{eq:grad}
\end{equation}
with  \(err_t^{m}:= \mathbb{I}[Y_t^{true} \notin C_{\alpha_t^m}^m] = 1\) if the predicted set does not contain the true label \(Y_t^{true}\), and $0$ otherwise. According to \eqref{eq:update-rule}, the adaptive miscoverage probability is increased when the prediction set includes the true label. This allows the prediction set to become smaller by excluding unnecessary labels \(\mathcal{Y}':= \{Y' \in \mathcal{Y} \mid \hat{q}_{\Bar{\alpha}_t^m}^m < S^{m}(X_t, Y') \le \hat{q}_{\alpha_t^m}^m\}\) in next step. Conversely, if the prediction set fails to include the true label, the adaptive miscoverage probability is decreased. \par
Additionally, the weights \(w_t^m\) for \(m \in S_t\) are updated after observing the true label \(Y_t^{true}\) by leveraging a multiplicative update rule:
\begin{equation}
w_{t+1}^{m} = w_t^{m} \exp\left(-\epsilon l_{t}^m/2^b\right),
\label{eq:update-rule-model-weight}
\end{equation}
where \(\epsilon\) is the step size that controls weight update, \(b = \lfloor \log_2 J \rfloor\) and \(l_t^m\) denotes the importance sampling loss estimates \cite{b4}
\begin{equation}
l_t^m = \frac{L\left(\Bar{\alpha}_t^m, \alpha_t^m\right)}{q_t^m} \mathbb{I}\{m \in S_t\},
\label{eq:lossfor_mocpgf}
\end{equation}
where \(q_t^m\) denotes the probability that the learning model \(m\) is included in  \(S_t\), which depends on how the graph \(G_t\) is generated. \par
Then, a weight \(u_{t+1}^j\) is assigned to each selective node \(j \in [J]\) according to the model nodes’ weights \(w_{t+1}^m\). Specifically, \(u_{t+1}^j\) is calculated as the sum of the weights \(w_{t+1}^m\) of all model nodes connected to the selective node \(j\), as follows:
    \begin{equation}
        \label{eq:u_jt formula}
        u_{t+1}^j = \sum_{\forall m: v_m^{(l)} \rightarrow  \in v_j^{(s)}} w^m_{t+1}.
    \end{equation}
 Moreover, the probability according to which a selective node is chosen in the next time step, denoted by \({p'}^j_{t+1}\), can be updated as \({p'}^j_{t+1} = \frac{u^j_{t+1}}{\sum_{i=1}^J u^i_{t+1}}\). To sum up, at each time step, all model nodes connected to the selected selective node form a subset of candidate learning models. This approach aims to avoid including low-performing models in the candidate set for the conformal prediction task. One model is then selected from this subset to construct the prediction set.

\subsection{Graph Generation}
This section describes the procedure for generating the graph \(G_t\). Let \(A_t\) represent the \(M \times J\) sub-adjacency matrix between two disjoint subsets \(\{v_1^{(l)}, ...,v_M^{(l)} \}\) and \(\{v_1^{(s)}, ...,v_J^{(s)} \}\). The entry \(A_t (m,j)\) denotes the \(m-\)th row and \(j-\)th column of matrix \(A_t\), and it's value is \(1\) if there is edge between model node \(m\) and selective node \(j\) in bipartite graph \(G_t\); otherwise it is \(0\). The probability of connecting model node \(v_m^{(l)}\) to each selective node is denoted by \(p_t^m\) and can be obtained as:
    \begin{equation}
        \label{eq:p_t^m formula}
        p_t^{m} = (1 - \eta_e)\frac{w_t^{m}}{\sum_{\Bar{m}=1}^{M} w_t^{\Bar{m}}} + \frac{\eta_e}{M}.
    \end{equation}
The second term in equation \eqref{eq:p_t^m formula}  introduces exploration among all model nodes. Specifically, each model node is connected to a selective node \(v_j^{(s)}\) uniformly at random if \(\eta_e = 1\). Each selective node \(v_j^{(s)}\) draws model nodes in \(N\) independent trials. In each trial, the selective node draws one model node according to PMF \(\boldsymbol{p}_t = (p_t^{m})_{m=1}^M\). According to the definition of \(p_t^m\) in \eqref{eq:p_t^m formula}, the probability that the \(m-\)th model node is connected to the \(j-\)th selective node is \(1- (1 - p_t^{m})^N\), where \((1 - p_t^{m})^N\) represents the probability that \(m\)th model node  is not selected by \(j-\)th selective node in any of \(N\) trials. Hence, the probability that the learning model \(m\) is included in \(S_t\) is given by
\begin{equation}
    q_t^m := \sum_{j=1}^J {p'}_t^j \left( 1 - \left( 1 - p_t^{m}\right)^N \right),
\end{equation}
for all \( m \in M\). This probability is used in the importance-sampling loss estimate defined in \eqref{eq:lossfor_mocpgf}. The entire process for generating the graph \(G_t\) is detailed in Algorithm \ref{alg:graph_generation}.

\RestyleAlgo{ruled}
\SetKwInput{KwRequire}{Require}
\SetKwInput{KwInitialize}{Initialize}
\SetKwFor{For}{for}{do}{end}

\begin{algorithm}
\caption{Generating Graph \(G_t\)}\label{alg:graph_generation}
\KwRequire{\(J\), \(N\), \(M\) pre-trained models.}
\KwInitialize{\(A_t = 0_{J\times M}\)}
Set \(p_t^{m} = (1 - \eta_e)\frac{w_t^{m}}{\sum_{\Bar{m}=1}^{M} w_t^{\Bar{m}}} + \frac{\eta_e}{M}, \forall m \in [M]\) \;
$X \gets x$\;
$N \gets n$\;
\For{\(j = 1, ...,J\)}{
  \For{\(n = 1, ..., N\)}{
  Select one of the models according to PMF \(\boldsymbol{p}_t = (p_t^{m})_{m=1}^M\) \;
  Set \(A_t(j, \Tilde{m}) = 1 \{\Tilde{m}\) is selected model from PMF\}\;
  }
}
\end{algorithm}

Given the graph \(G_t\) at each time step, one selective node is chosen according to the PMF \(\boldsymbol{p'}_t = ({p'}^j_t)_{j=1}^J\),  where \({p'}^j_t = \frac{u^j_t}{\sum_{i=1}^J u^i_t}\). The model nodes connected to the selected selective node form the candidate set. A single learning model is then selected from this set according to the PMF \(\boldsymbol{w}_t^s = (w_t^{\Bar{m}})_{\Bar{m} \in S_t}\) to construct the prediction set. The entire GMOCP method is summarized in Algorithm \ref{alg:GMOCP}.

\begin{algorithm}
\caption{Graph-Structured feedback Multi-model Ensemble Online Conformal Prediction (GMOCP)}\label{alg:GMOCP}
\KwRequire{\(\alpha \in [0,1]\), \(M\) models, and step size \(\epsilon \in (0, 1)\)}
\For{$t \in [T]$}{
    Receive new datum \(x_t\)\;
    Generate graph \(G_t\) using Algorithm 1\;
    Obtain \(u_t^j = \sum_{m \in v_j} w^m_t, \forall j \in [J]\)\;
  \For{\(j = 1, ..., J\)}{
  Set \({p'}^j_t = \frac{u^j_t}{\sum_{i=1}^J u^i_t}\)
  }
       Select one of the selective nodes according to the PMF \(\boldsymbol{p'}_t = ({p'}^j_t)_{j=1}^J\)\;
     Create a set \(S_t\) including connected models to the selected node.\;
     Obtain normalized weights by \(\Bar{w}_t^{m} = \frac{w_t^{m}}{\sum_{\Bar{m} \in S_t} w_t^{\Bar{m}}}, \forall m \in S_t\)\;
     Select model \(\hat{m}\) according to the PMF \(\boldsymbol{w}_t^s = (w_t^{\Bar{m}})_{\Bar{m} \in S_t}\)\;
     Obtain threshold \(\hat{q}^{\hat{m}}_{\alpha_t^{\hat{m}}}\) according to \eqref{eq:quantile_formula}, and construct prediction set  \( C_{\alpha_t^{\hat{m}}}^{\hat{m}} (X_t)\) via \eqref{eq:Create_pred_set}\;
     Observe the true label\;
     Calculates \(l_t^{\Bar{m}}\) and update \(w_t^{\bar{m}}\) and \(\alpha_t^{\bar{m}}\) according to \eqref{eq:lossfor_mocpgf}, \eqref{eq:update-rule-model-weight}, and \eqref{eq:update-rule}  \(\forall \Bar{m} \in S_t\) \;
}
\end{algorithm}

\section{Simulation Results}
This section evaluates the proposed algorithm, GMOCP, and demonstrates that it produces more efficient prediction sets while maintaining the desired coverage. We begin by explaining the experimental settings, and then compare the performance of our proposed method with a multi-model conformal prediction algorithm.  Note that throughout the experiments in this section, the desired miscoverage probability \(\alpha\) is \(0.1\). All experiments were performed on a workstation with NVIDIA RTX A4000 GPU.

\subsection{Datset:} We conduct experiments on both synthetic and real datasets. For the synthetic setting, we simulate online environments using two distinct transformation sequences. From each transformation sequence, two datasets are generated with random variations to ensure diversity across samples. Each synthetic dataset contains 3,000 images spanning 20 classes. Gradual distribution shifts are modeled by sampling within a single transformation type, whereas abrupt shifts are created by alternating between datasets derived from different transformations. For the real-data experiments, we use TinyImageNet-C, a corrupted variant of the TinyImageNet dataset containing 200 classes.

\subsection{Score Function:} The nonconformity score defined in \cite{b5} is utilized to construct prediction sets. Let
\begin{equation}
   S^{m}(X,Y) = \xi \sqrt{\max([k_Y - k_{reg}], 0)} + U_t\hat{f}_Y^{m}(X) + \rho (X,Y),
  \label{eq:nonconformity score}   
\end{equation}
where \( \hat{f}_Y^{m}(X) \) denotes the probability of predicting label \( Y \) for input \( X \) by model \( m \), and \( U_t \) is a random variable sampled from a uniform distribution over the interval \([0, 1]\). The term \( k_Y:= |\{Y' \in \mathcal{Y} \mid  \hat{f}_{Y'}^{m}(X) \geq \hat{f}_Y^{m}(X) \}|\) denotes the number of labels that have a higher or equal predicted probability than label \( Y \) according to the model’s output probability distribution, e.g., the softmax output. \( \rho (X, Y):= \sum_{Y'=1}^{N_{\text{labels}}} \hat{f}_{Y'}^{m}(X) \mathbb{I}[\hat{f}_{Y'}^{m}(X) > \hat{f}_Y^{m}(X)] \) sums up the probabilities of all labels that have a higher predicted probability than label \( Y \).

\subsection{Evaluation Metrics:} Coverage measures the percentage of instances in which the true label is included in the prediction sets constructed by the conformal prediction algorithm over the period $[T]$. Avg Width represents the average size of the prediction sets constructed from \(t=1\) to \(T\). Run Time indicates the time required to complete the algorithm for one random seed.

\subsection{Learning Models:}We employ \(6\) candidate learning models: GoogLeNet \cite{b7}, ResNet-50, ResNet-18 \cite{b8}, DenseNet121 \cite{b9}, MobileNetV2 \cite{b10}, and EfficientNet-B0 \cite{b11}. To ensure a diverse range of performance across these models, each one is trained under \(3\) distinct settings: High-performance setting (the model is trained for 120 epochs and initialized with default pretrained weights from ImageNet \cite{b12}), Medium-performance setting (the model is trained for only 10 epochs and initialized with random weights, resulting in weaker performance), Low-performance setting (the model is trained for just 1 epoch with random weight initialization, yielding the weakest performance among the \(3\)). For clarity, each model is labeled according to both its architecture and training configuration. For example, the three DenseNet121 variants are denoted as DenseNet121-120D (120 epochs, pretrained weights), DenseNet121-10N (10 epochs, random initialization), and DenseNet121-1N (1 epoch, random initialization). In all \(3\) settings, the learning rate is set to \(10^{-3}\), and the batch size is fixed at \(64\).

\begin{table}[htbp]
\centering
\caption{
Results on the Synthetic dataset, evaluated across different values of $N$ and $J$.
The target coverage is 90\%. Bold numbers denote the best results in each column.
}
\label{tab:synthetic_two_column}
\begin{tabular}{|c | c | l | c | c | c|}
\hline
\textbf{N} & \textbf{J} & \textbf{Method} &
\textbf{Coverage (\%)} &
\textbf{Avg Width} &
\textbf{Run Time} \\
\hline
 &  & MOCP   & \textbf{89.92 $\pm$ 0.28} & 18.00 $\pm$ 0.01 & 13.57 $\pm$ 0.18 \\
\hline
1 & 1 & GMOCP   & 89.24 $\pm$ 0.12 & 17.88 $\pm$ 0.06 & \textbf{5.96 $\pm$ 0.05}  \\
  & 2  &  GMOCP  & 89.26 $\pm$ 0.15 & \textbf{17.86 $\pm$ 0.07} & 6.17 $\pm$ 0.05  \\
\hline
3 & 1 &  GMOCP  & 89.60 $\pm$ 0.20 & 17.96 $\pm$ 0.04 & 7.17 $\pm$ 0.27  \\
 & 2 &  GMOCP  & 89.78 $\pm$ 0.12 & 17.97 $\pm$ 0.04 & 8.90 $\pm$ 0.14 \\
\hline
5 & 1 &  GMOCP  & 89.77 $\pm$ 0.27 & 17.95 $\pm$ 0.02 & 7.88 $\pm$ 0.45  \\
 & 2 &  GMOCP  & 89.73 $\pm$ 0.26 & 17.97 $\pm$ 0.04 & 8.90 $\pm$ 0.14  \\
  
\hline
\end{tabular}
\end{table}

\subsection{Baseline:} The proposed method is compared with MOCP \cite{b6}  algortihm. The MOCP algorithm employs \(M\) learning models and selects one model from the entire set at each time step \(t\). The selection is based on the weights assigned to each model, and the prediction set is constructed using the selected model.
\subsection{Results:} For this section, experiments are conducted using a candidate set of eight different learning models, including: DenseNet121-120D, ResNet-18-120D, GoogLeNet-120D, ResNet-50-120D, MobileNetV2-120D, EfficientNet-B0-120D, DenseNet121-10R, and DenseNet121-1R. For synthetic datasets, we evaluate the performance of GMOCP under different settings for \(N\) and \(J\). Table \ref{tab:synthetic_two_column} demonstrates that GMOCP is able to obtain smaller prediction sets while preserving the desired coverage. Moreover, GMOCP achieves a lower computational cost than MOCP and results in shorter runtime across all evaluated configurations. \par
To further validate the method on real data, we conduct experiments on the TinyImageNet-C dataset, and the results are presented in \ref{tab:tiny_image}. The results show that GMOCP
consistently obtains smaller prediction sets in less time compared to MOCP.

\begin{table}[htbp]
\centering
\caption{
Results on the Synthetic dataset, evaluated across different values of $N$ and $J$. The target coverage is \(90\%\). Bold numbers denote the best results in each column.
}
\label{tab:tiny_image}
\begin{tabular}{|c | c | l | c | c | c|}
\hline
\textbf{N} & \textbf{J} & \textbf{Method} &
\textbf{Coverage (\%)} &
\textbf{Avg Width} &
\textbf{Run Time} \\
\hline
 &  & MOCP   & \textbf{89.61 $\pm$ 0.46} & 170.59 $\pm$ 1.03 & 3.01 $\pm$ 0.01\\
\hline
1 & 1 & GMOCP   & 87.90 $\pm$ 0.28 & 165.68 $\pm$ 1.34 & \textbf{2.03 $\pm$ 0.01}  \\
  & 2  &  GMOCP  & 87.81 $\pm$ 0.40 & \textbf{165.57 $\pm$ 2.08} & 2.12 $\pm$ 0.03  \\
   & 4 &  GMOCP  & 87.54 $\pm$ 0.27 & 165.28 $\pm$ 1.77 & 2.26 $\pm$ 0.01 \\
\hline
3 & 1 &  GMOCP  & 88.91 $\pm$ 0.37 & 167.66 $\pm$ 1.43 & 2.38 $\pm$ 0.02  \\
 & 2 &  GMOCP  & 88.98 $\pm$ 0.38 & 168.01 $\pm$ 1.29 & 2.56 $\pm$ 0.03 \\
  & 4 &  GMOCP  & 88.98 $\pm$ 0.34 & 168.98 $\pm$ 1.01 & 2.85 $\pm$ 0.02 \\
\hline
\end{tabular}
\end{table}

\section*{CONCLUSION}

The present paper introduced an algorithmic framework for online conformal prediction with multiple learning models, where a subset of candidate models must be selected to construct prediction sets. By constructing a graph that reflects model performance over previous time steps, we proposed a data-driven mechanism for selecting an effective subset of models. This approach leads to smaller prediction sets and reduced computational complexity while preserving the desired coverage. Experimental results demonstrate that the proposed method outperforms the multi-model conformal prediction algorithm in both prediction set size and computational complexity.

\bibliographystyle{IEEEtran}
\bibliography{main}

\end{document}